\documentclass[sigconf]{acmart}

\usepackage[linesnumbered,ruled,vlined]{algorithm2e}

\usepackage{listings}
\usepackage{dsfont}

\SetCommentSty{mycommfont}

\SetKwInput{KwInput}{Input}                
\SetKwInput{KwOutput}{Output}              

\AtBeginDocument{%
  }

\copyrightyear{2023}
\acmYear{2023}
\setcopyright{acmlicensed}\acmConference[SIGIR '23]{Proceedings of the 46th International ACM SIGIR Conference on Research and Development in Information Retrieval}{July 23--27, 2023}{Taipei, Taiwan}
\acmBooktitle{Proceedings of the 46th International ACM SIGIR Conference on Research and Development in Information Retrieval (SIGIR '23), July 23--27, 2023, Taipei, Taiwan}
\acmPrice{15.00}
\acmDOI{10.1145/3539618.3591908}
\acmISBN{978-1-4503-9408-6/23/07}

\begin{document}

\title{RiverText: A Python Library for Training and Evaluating Incremental Word Embeddings from Text Data Streams}



\author{Gabriel Iturra-Bocaz}
\affiliation{%
 \small{  \institution{Department of Computer Science, University of Chile}
  \institution{National Center for Artificial Intelligence (CENIA)}
  \institution{Millennium Institute for Foundational Research on Data (IMFD)}
  \city{Santiago}
  \country{Chile}
}}
\email{gabrieliturrab@ug.uchile.cl}

\author{Felipe Bravo-Marquez}
\affiliation{%
 \small{  \institution{Department of Computer Science, University of Chile}
  \institution{National Center for Artificial Intelligence (CENIA)}
  \institution{Millennium Institute for Foundational Research on Data (IMFD)}
  \city{Santiago}
  \country{Chile}
}}
\email{fbravo@dcc.uchile.cl}

\renewcommand{\shortauthors}{Iturra-Bocaz \& Bravo-Marquez}

\begin{abstract}

Word embeddings have become essential components in various information retrieval and natural language processing tasks, such as ranking, document classification, and question answering. However, despite their widespread use, traditional word embedding models present a limitation in their static nature, which hampers their ability to adapt to the constantly evolving language patterns that emerge in sources such as social media and the web (e.g., new hashtags or brand names). To overcome this problem, incremental word embedding algorithms are introduced, capable of dynamically updating word representations in response to new language patterns and processing continuous data streams.

This paper presents RiverText, a Python library for training and evaluating incremental word embeddings from text data streams. Our tool is a resource for the information retrieval and natural language processing communities that work with word embeddings in streaming scenarios, such as analyzing social media. The library implements different incremental word embedding techniques, such as Skip-gram, Continuous Bag of Words, and Word Context Matrix, in a standardized framework. In addition, it uses PyTorch as its backend for neural network training.

We have implemented a module that adapts existing intrinsic static word embedding evaluation tasks for word similarity and word categorization to a streaming setting. Finally, we compare the implemented methods with different hyperparameter settings and discuss the results.  

Our open-source library is available at \url{ https://github.com/dccuchile/rivertext}.

\end{abstract}




\maketitle

\section{Introduction}\label{sec:intro}

Word embeddings (WE) \cite{mikolov2013distributed,turney2010frequency}, which are mappings from discrete words to dense continuous vectors, have been widely incorporated into modern information retrieval (IR) and natural language processing (NLP) systems because of their inherent ability to capture the syntactic and semantic properties of words. 

There are essentially two main approaches to constructing these vectors (which are equivalent in \cite{levy2014neural}): 1) count-based approaches that explicitly build a word-context matrix of word-word co-occurrence counts \cite{turney2010frequency} and 2) distributed methods that rely on the internal structure of a neural network trained on an auxiliary predictive task (e.g., predicting the center word from a context window) \cite{mikolov2013distributed}. 

However, the static nature of standard WE algorithms prevents them from incorporating new words, such as hashtags or new brand names, and adapting to semantic changes in existing words. For example, when unexpected events associated with an entity suddenly occur (e.g., a scandal related to a public figure or a company may change its perceived sentiment). Another scenario is when a given word acquires a new meaning in a particular event; for example, the word ``ukrop'' changed its meaning from ``dill'' to ``Ukrainian patriot'' during the Russian-Ukrainian crisis \cite{stewart2017measuring}.  To incorporate these changes into traditional WE, they must be retrained or aligned with new models to incorporate knowledge from new text sources, which is computationally inefficient.  

The research community has proposed several methods for incremental learning in WE to address the drawbacks in traditional approaches. However, these methods, such as the incremental word context matrix (IWCM) \cite{bravo2022incremental}, incremental skip-gram negative sampling (ISG) \cite{kaji2017incremental,may2017streaming}, and incremental hierarchical softmax \cite{peng2018incremental}, lack a unified and transparent setup for comparison, hindering the examination and understanding of their quality and performance. This lack of information is crucial for deploying these algorithms in real-world NLP or IR systems.

In this paper, we integrate these attempts into a new Python library called RiverText. This resource aims to unify and standardize the aforementioned methods into an easy-to-use toolkit. RiverText extends the interfaces provided by River \cite{montiel2021river}, a machine-learning library for data streams, by enabling continuous learning of word embeddings from text streams, either from one instance at a time or from a mini-batch of \cite{readbatch} examples. In addition, it uses PyTorch \cite{paszke2019pytorch} as its backend for implementing neural networks.

RiverText (as well as River) is based on the stream machine learning paradigm \cite{geng2009incremental, readbatch, ade2013methods}, which posits the following requirements for a learning algorithm \cite{aggarwal2007data, muthukrishnan2005data,bifet2018machine}:
\begin{enumerate}
    \item be able to process one instance (or mini-batch) at a time, and inspect it (at most) once;
    \item be able to process data with limited resources (time and memory);
    \item be able to generate a prediction or transformation at any time;
    \item be able to adapt to temporal changes.
\end{enumerate}

In RiverText, we developed a standardized procedure to evaluate incremental WE methods in order to track the quality of embeddings throughout the stream. Our procedure performs a periodic evaluation (e.g., after processing a certain number of text instances) of existing intrinsic WE evaluation tasks for word similarity and categorization.  In addition, we perform a comprehensive evaluation of three incremental WE methods: IWCM, ISG, and incremental continuous bag-of-words (ICBOW). 

We list our contributions below: 

\begin{itemize}
    \item To the best of our knowledge, we develop the first open-source library that standardizes the training and evaluation of incremental WE from text data streams.
    \item We propose the first implementation of ICBOW with negative sampling that combines the techniques exposed in the ISG method.
    \item We perform the first unified benchmarking procedure of several incremental WE models.
\end{itemize}

This article is organized as follows. First, in Section~\ref{sec:related}, we review related work. Second, in Section~\ref{sec:software}, we explain an overview of the software. In Section~\ref{sec:experiments}, we present our experiments showing our benchmark. The conclusion and future work are discussed in Section~\ref{sec:conc}.


\section{Related Work}\label{sec:related}

In this section, we review the literature on the three main aspects on which this work is based: 1) incremental WE models, 2) stream machine learning libraries, and 3) intrinsic evaluation of WE. First, we cover the models implemented in our framework and others, such as Incremental GloVe \cite{pennington2014glove}, which is not added to our library but will be included in the next version. Second, we review the main libraries used by the research community for machine learning of data streams. Finally, we discuss intrinsic evaluation approaches for WE.

\subsection{Incremental Word Embedding Models}

As we pointed out previously, WE can be divided into count-based approaches and distributed methods, both based on the distributional hypothesis \cite{harris1954distributional} (i.e., words appearing in the same contexts tend to have similar meanings). According to this classification, we implemented the following models discussed below:

The Incremental Word Context Matrix model (proposed by Bravo-Marquez et al. \cite{bravo2022incremental}) is a count-based method that constructs a word-context matrix of size $V \times C$, where $V$ is the number of words contained in the vocabulary and $C$ is the number of contexts around the target words (obtained from a surrounding window of fixed size). Each matrix cell encodes the association between a target word and a context, computed  using a smoothed positive point-wise mutual information (PPMI) score \cite{martin2009speech}. To keep memory usage constant throughout  the stream, it is necessary to keep the number of words fixed, composing the vocabulary, the contexts, and the counters calculating the PPMI weights. When there is no space for a new word or context, the existing ones are replaced according to a given criterion (less frequent word/context, older word/context, among others).

Incremental skip-gram with negative sampling (ISG) is based on the neural network architecture proposed by Kaji and Kobayashi \cite{kaji2017incremental}. This model is inspired by the original skip-gram from the Word2Vec library (Mikolov et al. \cite{mikolov2013distributed}). Kaji and Kobayashi develop an incremental version of negative sampling; their algorithm builds a unigram table that incrementally updates the words frequencies and the noise distribution. The authors use the Misra Gries algorithm \cite{misra1982finding} to allocate words dynamically in a finite vocabulary using constant memory throughout the stream.

Other methods not implemented yet, but to be added in the future are:

SpaceSaving Word2Vec (SSW) is a work similar to ISG but developed independently by May et al. \cite{may2017streaming}. The main differences are:

\begin{itemize}
    \item ISG uses the Misra Greis algorithm \cite{misra1982finding} for dynamic word allocation, while SWE employs the Space Save algorithm \cite{metwally2005efficient}, which counts the most frequent elements in a data stream.
    \item SSW uses the original unigram sampling table to estimate the negative distribution; Kaji and Kobadashi proposed an original algorithm \cite{kaji2017incremental} for this purpose.
\end{itemize}

 The incremental Glove \cite{peng2017incrementally} model follows the same idea as the original GloVe \cite{pennington2014glove}; calculates the global statistics with a word-context matrix of co-occurrences, and reduces the matrix dimensionality by training a square-root loss function. The main difference is that the incremental version modifies the loss function into a recursive scheme that depends on old and new data.

\subsection{Stream Machine Learning Libraries}
As discussed above, in the stream machine learning setting, models learn continuously from data streams that evolve over time. This learning can typically be done in two ways: 1) training one example at a time or 2) training by mini-batches of examples. An important difference with the standard machine learning paradigm is that stream models cannot perform data preprocessing operations that require full access to the data (e.g., vocabulary extraction).  Note also that stream machine learning is very similar to the incremental or online learning paradigms in machine learning, but incorporates some additional constraints, such as those listed in Section~\ref{sec:intro}. 

Massive Online Analysis (MOA), developed by Bifet et al. \cite{bifet2010moa}, is a Java software package that implements numerous machine learning algorithms for training and evaluation from evolving data streams. In addition, Bifet et al. developed MOATweetReader \cite{bifet2011moa}, an extension to MOA for analyzing tweets in real time, detecting changes in word distribution, performing summary statistics, and sentiment analysis.

River \cite{montiel2021river} was formed from the union of two similar predecessor projects, Creme \cite{halford2019creme}, and scikit-multiflow \cite{montiel2018scikit}, which provides Python implementations of the main machine learning algorithms for data streams for tasks such as classification, regression, and clustering, as well as other functionalities. In standard machine learning, multidimensional arrays are typically used as the primary data structure for data representation. However, since streaming data can come up at any time, River uses dictionaries as a more flexible and faster alternative. To optimize mathematical operations between dictionaries, River relies on its own dictionary data structure, called \texttt{VectorDict}, which is implemented in Cython \cite{behnel2010cython}.

Note that none of these libraries are designed to perform representation learning for unstructured data, such as word embeddings, in an incremental fashion.

\subsection{Intrinsic Evaluation}

The WE intrinsic evaluation is a family of evaluation techniques for measuring the syntactic and semantic properties captured by these vectors that include three types of tasks: word similarity (i.e., whether the similarity between two words vectors correlates with a human judgment of relatedness), analogies (i.e., when relations in the form of ``a is to b as c is to d'' can be obtained from arithmetic operations on the vectors), and categorization (i.e., when groups of words are aligned with predefined categories, such as animals).
These evaluations are often combined to benchmark different WE algorithms, the corpora on which they are trained, and the hyperparameter settings \cite{schnabel2015evaluation}.  The Word Embeddings Benchmark\footnote{https://github.com/kudkudak/word-embeddings-benchmarks} \cite{jastrzebski2017evaluate} is an open-source package that brings together all intrinsic evaluations into a unified interface to facilitate the evaluation and comparison of these resources. However, this evaluation approach has its detractors, Gladkova and Drozd \cite{gladkova2016intrinsic} argued that intrinsic evaluation ignores key features of distributional semantics (e.g., polysemy), and does not always correctly determine how a word embedding would perform in a downstream application.

It is important to note that these evaluations are designed for a standard machine learning setting (i.e., the evaluation is performed after the training is completed). In this work, we attempt to adapt them to a streaming setting.

\section{Software Overview}\label{sec:software}

RiverText is a Python library for training and evaluating incremental WE architectures from text data streams. Our library extends the River base class, \texttt{Transformer}, where it implements the methods \texttt{learn\_one} and \texttt{learn\_many}, which refer to training one example at a time or from  mini-batches of examples, respectively.

As mentioned in previous sections, incremental WE training is the process of learning dynamic word vectors from continuously arriving streams of text, such as tweets. The process in which these vectors are trained in our software is as follows:

\begin{enumerate}
    \item Connect to a continuous source of a text data stream (e.g., Twitter).
    \item Tokenize the text and traverse its words.
\item  If a new word is found, it is added to the vocabulary and a new vector is assigned to it.
    \item If the word is known, its corresponding vector is updated according to its context (i.e., its surrounding words).
    \item At any time during training, it is possible to get the vector associated with a vocabulary word.
\end{enumerate}

RiverText users can modify the incremental word embedding learning algorithm (depending on available implementations) and select a sketching algorithm to dynamically allocate new words to the vocabulary. This is an important component of our framework since it is impossible in a streaming setting to determine the vocabulary size in advance, as it is usually done in standard machine learning. The current version of the software only implements Misra Gries' algorithm \cite{misra1982finding}, but we plan to add more algorithms in the future, such as the Space Saving algorithm \cite{metwally2005efficient}.

In the following subsections, we will describe the incremental learning approaches implemented in our software (instance incremental and batch incremental), our periodic evaluation procedure, the learning methods implemented, and a general overview of our software’s architecture.

\subsection{Incremental Learning Approaches}\label{ssec:inc}

RiverText implements two incremental learning approaches: 1) instance incremental, and 2) batch incremental learning, which are discussed below.

In Instance Incremental learning, our WE parameters are updated with every training instance (e.g., a tweet) and discarded after training, as shown below:

\lstset{language=Python}
\lstset{frame=lines}
\lstset{caption={Example of \texttt{learn\_one} method using the Incremental WCM model. The parameters of the \texttt{WordContextMatrix} class are the vocab size, window size, and context size, respectively.}}
\lstset{label={lst:learn_one}}
\lstset{basicstyle=\footnotesize}
\begin{lstlisting}
from rivertext.models import WordContextMatrix
from rivertext.utils import TweetStream

from torch.utils.data import DataLoader

ts = TweetStream("/path/to/tweets.txt")
wcm = WordContextMatrix(
    vocab_size=100000, 
    window_size=3, 
    context_size=1000
)
dataloader = DataLoader(ts, batch_size=1)

for tweet in dataloader:
    wcm.learn_one(tweet)
\end{lstlisting}

 In this case, the text stream is simulated from a file of tweets (one tweet per line and separated by a broken line) and read from the buffer one at a time using PyTorch \texttt{DataLoader} with a batch size of 1. Then, our learning algorithm (IWCM in this case) only has to call the \texttt{learn\_one} method to update its parameters accordingly.

This approach suffers from efficiency problems due to the overhead of processing one instance at a time. In addition, in the case of neural network-based models, such as ISG and ICBOW, which are based on gradient descent, loss calculations can result in inaccurate gradients \cite{goldberg2017neural}, which can lead to requiring too many instances to obtain good word representations.

Incremental batch learning, on the other hand, gathers a small batch of instances before training. This allows neural network-based models to benefit from the increased efficiency of specialized computing architectures such as GPUs, which replace vector-matrix operations with matrix-matrix operations for forward and backward network passes. The difference with traditional batch learning is that batches can only be processed once and must be deleted once processed.

\lstset{language=Python}
\lstset{frame=lines}
\lstset{caption={Example of \texttt{learn\_many} method using the Incremental WCM model. The parameters of the \texttt{WordContextMatrix} class are the vocab size, window size, and context size, respectively.}}
\lstset{label={lst:learn_many}}
\lstset{basicstyle=\footnotesize}
\begin{lstlisting}
from rivertext.models import WordContextMatrix
from rivertext.utils import TweetStream

from torch.utils.data import DataLoader

ts = TweetStream("/path/to/tweets.txt")
wcm = WordContextMatrix(
    vocab_size=100000, 
    window_size=3, 
    context_size=1000
)
dataloader = DataLoader(ts, batch_size=32)

for batch in dataloader:
    wcm.learn_many(batch)
\end{lstlisting}

As shown in Listing \ref{lst:learn_many}, our learning models only need to call the \texttt{learn\_many} method to process and train a batch of instances. The text stream is also read in batches of $w$ tweets using PyTorch's \texttt{DataLoader} with a batch size of $w$. 

An appropriate batch of $w$ usually depends on the available GPU memory capacity. It is important to note that with this approach, the word vectors will not be updated until the batch has been processed.

\subsection{Periodic Evaluation}\label{ssec:pe}

The proposed method for evaluating our incremental WE performance is called Periodic Evaluation. This method applies a series of evaluations to the entire model, using a test dataset associated with intrinsic NLP tasks after a fixed number, $p$, of instances, have been processed and trained. The algorithm takes as input the following arguments:

\begin{itemize}
    \item The parameter $p$ represents the number of instances between the evaluation series.
    \item The incremental WE model, referred to as $M$, is to be evaluated.
    \item The input text data stream, referred to as $TS$, used to train the incremental WE model.
    \item A test dataset, $GR$, associated with intrinsic NLP tasks.
\end{itemize}

The Periodic Evaluation algorithm aims to offer a structured evaluation scheme of an incremental word embedding model throughout the training process. It provides a mechanism for continuously assessing the model's performance, thereby enabling the identification of any potential issues and offering valuable insights into the model's progress. However, it should be noted that while traditional evaluation methods for NLP tasks have been applied in static settings, the Periodic Evaluation represents a novel approach by extending their functionality to the dynamic scenario of text streams, where the models can be trained indefinitely.

\begin{algorithm}[!ht]
\DontPrintSemicolon
  
  \KwInput{Stream ST, Incremental WE model, Intrinsic Dataset GR, int p}
    
    $c = 0$ \\
   \While{batch in ST}
   {
        \tcp{train the model}
   		learn\_many(model, batch) \\
        \tcp{evaluate the model during certain periods}
        \If{$c \neq 0 \wedge c \ mod \ p$} {
            result = evaluator(model.wv, GR) \\
            \tcp{the result is stored in a JSON file} 
            save(result) \\
        }
        c += length(batch) \\
   }
\caption{Periodic Evaluation Algorithm. The evaluator function takes the words and their mapped vectors, and an intrinsic dataset.}
\label{alg:period}
\end{algorithm}

In Algorithm \ref{alg:period}, we can observe how the periodic evaluation is implemented. In line 5, there is a function referred to as ``evaluator,'' which takes as input the vocabulary structure, the mapped vectors, and the test dataset associated with intrinsic NLP tasks and reduces the quality of word embeddings formed into a scalar value. This function provides a quantitative measure of the quality of the word embeddings generated by the incremental WE model, allowing for a more accurate assessment of its performance.

The intrinsic tasks implemented in the proposed method are similarity, analogies, and categorization. The following evaluation metrics measure these tasks:
\begin{itemize}
    \item \textbf{Similarity}: The Spearman correlation coefficient \cite{wissler1905spearman}, denoted as $\rho$, is used to calculate the degree of association between the similarity scores calculated from the word embeddings and the scores obtained from a human-annotated dataset.
    \item \textbf{Analogies}: Accuracy is used to count the number of correctly obtained words from an analogy equation, comparing the set of analogy words obtained from the word embeddings with the set of analogy words from the human-annotated dataset.
    \item \textbf{Categorization}: Purity clustering \cite{manning2008introduction} is used to count the total number of correctly classified words, comparing the categories obtained from the word embeddings with the categories from the human-annotated dataset.
\end{itemize}

The results of the evaluation metrics are stored as a JSON file that the user can access and examine at any time.

We have delegated the implementation of the evaluator for intrinsic tasks to an external library called Word Embedding Benchmark \cite{jastrzebski2017evaluate}. This library provides a comprehensive collection of test datasets (e.g., MEN \cite{bruni2014multimodal}, MTURK \cite{radinsky2011word}, AP \cite{almuhareb2005concept})  and the corresponding methods for measuring the quality of word embeddings. The intrinsic task in question determines the specific evaluator function to be utilized. For instance, the similarity task requires using the \texttt{evaluate\_similarity} function provided by the library. For further information on this library's functionality and usage, refer to the GitHub repository\footnote{https://github.com/kudkudak/word-embeddings-benchmarks}.

It is important to note that the Periodic Evaluation method only assesses the quality of word embeddings for words present in the model's vocabulary at the evaluation time. As the vocabulary is subject to changes due to the application of the Misra-Gries algorithm for discarding infrequent words, the word embeddings for discarded words are not evaluated unless they are subsequently reintroduced to the vocabulary.

Another important consideration is if the evaluator's test dataset contains words that are not included in the model's vocabulary, the out-of-vocabulary words are assigned the average embedding of the words that are present in the vocabulary. This process is used to evaluate the quality of the model. It is crucial to note that this approach can impact the model's overall performance, and the results should be interpreted cautiously.

\subsection{Implemented Methods}

For this work, we adapted and modified three models: the Incremental Word Context Matrix (IWCM), Incremental SkipGram with Negative Sampling (ISG) and Incremental Continuous Bag of Words with Negative Sampling (ICBOW), whose details we explain next.

\subsubsection{Incremental Word Context Matrix}

Our implementation of the IWCM model is based on the algorithm described in the work of Bravo-Marquez et al. \cite{bravo2022incremental}. The IWCM model utilizes a co-occurrence matrix of dimension $V$ x $C$, where $V$ represents the number of words present in the vocabulary and $C$ represents the number of context words associated with each target word. It is essential to note that, as opposed to its static counterpart, the co-occurrence matrix in the IWCM model may not be square due to the incremental nature of the algorithm. The relationship between a target word and its context is weighted by the Positive Pointwise Mutual Information (PPMI) score \cite{daniel2007speech}, a commonly used measure of association in NLP.

\begin{equation} \label{eq:pmi}
PPMI(w, c) =  max\left(0, log_2 \left( \frac{count(w, c) \times D}{count(w) \times count(c)} \right)\right) 
\end{equation}

In Equation \ref{eq:pmi}, the variable $D$ represents the total number of words in the text streams. The counters, $count(w_i, c_j)$, $count(w_i)$, $count(c_j)$, and $D$, which are used to calculate the probabilities of the word-context pairs for the PPMI score, are efficiently stored in \texttt{VectorDict} objects provided by the River packages \cite{montiel2021river}. These objects function as a sparse data structure, enabling efficient mathematical operations and incremental updates of the word-context matrix.

\begin{algorithm}[!ht]
\DontPrintSemicolon
  
  \KwInput{Stream ST, window size W, vocab size V, context size C}
  \KwOutput{Matrix Mat V x C}
    $d = 0$ \\
   \While{batch in ST}
   {
        \For{tweet in batch}    
        { 
        	tokens = tokenize(tweet) \\
            \For{token in tokens} {
                \If{token not in vocab} {
                    addToVocab(vocab, token)
                }
                d += 1 \\
                contexts = getContexts(token, tokens, W) \\
                updateDictCounter(token, contexts) \\
                \For{cont in contexts} {
                    Mat(token, cont) = $max \left(0, log \left(\frac{count(token, cont) \cdot d}{count(token) \cdot count(cont)}\right) \right)$
                }
            }
            \tcp{reduce the embedding dimension by incremental PCA}
            reduceEmbDimByIPCA(tokens) \\
        }
   }

\caption{IWCM model method.}
\label{alg:wcm}
\end{algorithm}

In Algorithm \ref{alg:wcm}, a sliding window of 2W tokens is utilized to extract context information from the tokenized tweets in the text stream. The center of the window is aligned with a target word, and all surrounding tokens within the window are considered context tokens. For unseen target words and contexts, new entries are dynamically allocated in the \texttt{VectorDict} objects and initialized with a count value of zero. For existing words and contexts, the corresponding counters are updated incrementally. This approach allows for efficient storage and manipulation of the word-context matrix while maintaining an acceptable level of accuracy.

One limitation of this method is that, similar to its static counterpart, the IWCM model produces sparse and high-dimensional vectors. To address this issue, we employ the incremental Principal Component Analysis (PCA) \cite{artac2002incremental} technique, to reduce the dimensionality of the generated embeddings. This algorithm does not require multiple passes over the entire set of embeddings to achieve dimensionality reduction, as it processes the data as a vector stream. In addition, our IWCM implementation selectively applies dimensionality reduction to recently added or updated embedding vectors to optimize computational efficiency.

\subsubsection{Incremental Word2Vec}

The incremental Word2Vec architecture comprises two ISG and ICBOW models based on the static version proposed by Mikolov et al. \cite{mikolov2013distributed}. The ISG model predicts the context words for a given target word, and the ICBOW model aims to predict the target word using its context words.

Our implementation is based on the Skip Gram model with Negative Sampling, as proposed by Kaji and Kobayashi \cite{kaji2017incremental}. This implementation extends the traditional unigram table, typically created as a static word array, to an incremental approach. Instead of performing multiple passes over the entire dataset to complete the unigram table, the model updates the table incrementally, making the process more efficient and scalable.

\begin{algorithm}[!ht]
\DontPrintSemicolon
  
  \KwInput{Array word\_indexes, Array T, int size\_T, Array Freqs, float $\alpha$}
  \KwOutput{Array T}
    $z = 0$ 

    \For{index in word\_indexes}    
    { 
        Freqs[index] += 1 \\
        $F = Freqs[index] - (Freqs[index] - 1)^\alpha$ \\
        z += F 
        
        \If{$|T| < size\_T$} {
            add F copies of index to T 
        } 
        \Else {
            \For{j = 1, ..., $\frac{size\_T \cdot F}{z}$}{
               T[j] = index with probability $\frac{F}{z}$
            } 
        }

    }
\caption{Adaptive Unigram Table.}
\label{alg:uni}
\end{algorithm}

In Algorithm \ref{alg:uni}, we present the adaptive unigram table proposed by Kaji and Kobayashi \cite{kaji2017incremental}. Given a fixed-size unigram table T with a capacity of $size\_T$, an array $Freqs$ representing the frequencies of the words in the vocabulary, a tuple of word indexes representing a tweet, and a smoother parameter $\alpha$. The algorithm proceeds as follows:

\begin{itemize}
    \item If the number of elements in $T$, $|T|$, is less than $size\_T$, $F$ copies of the word index are added to T, where the word index corresponds to the indexes mapped to the words that compose the vocabulary.
    \item Otherwise, the number of copies of the word index added to $T$ is calculated as $\frac{size\_T \cdot F}{z}$, and the new additions to $T$ may overwrite current values with a probability proportional to $\frac{F}{z}$. This process updates the distribution of words represented in $T$.
\end{itemize}

It is important to note that the frequency of each word is proportional to the number of its indexes stored in T.

\begin{algorithm}[!ht]
\DontPrintSemicolon
  
  \KwInput{Stream ST, Vocab size V, Unigram Table Size T,int num\_ns}
    
    vocab = Vocab(V) \\
    ut = UnigramTable(T) \\
   \While{batch in ST}
   {
        \For{tweet in batch}    
        { 
        	tokens = tokenize(tweet) \\
            \For{w in tokens} {
                \If{w not in vocab} {
                    addToVocab(vocab, w)
                }
                updateTokenFreq(w) \\
                updateUnigramTable(w) \\
                contexts = getContexts(w, tokens) \\
                \For{c in contexts} {
                    draw $num\_ns$ indexes from ut: $ns_{1}, ns_{2}, ..., ns_{num\_ns}$ \\
                    \tcp{convert the word indexes to the neural model input} 
                    $v_{w}$, $v_{c}$, $ns_{1}$, $ns_{2}$, $...$, $v_{ns_{num\_ns}}$ = preprocessing($w$, $c$, $ns_{1}$, $ns_{2}$, $...$, $v_{ns_{num\_ns}}$) \\
                    \tcp{performs SGD to update the word embedding} 
                    SGD($v_{w}$, $v_{c}$, $v_{ns_{1}}$, $v_{ns_{2}}$, $...$, $ns_{num\_ns}$) \\ 
                }
            }
        }
   }

\caption{Incremental Word2Vec method}
\label{alg:w2v}
\end{algorithm}

In Algorithm \ref{alg:w2v}, the adaptive unigram method is implemented through the functions \texttt{updateTokenFreq} and \texttt{updateUnigramTable}. The two incremental word2vec models utilize this algorithm: ISG and ICBOW, with the only difference being the neural architecture used. However, a crucial preprocessing step is necessary before performing the stochastic gradient descent in line 14. This step involves converting the word indexes into the appropriate input format for the specific ISG or ICBOW models and is essential for the proper functioning of the algorithm.

The RiverText package incorporates the implementation of the neural network backend for both the ISG and ICBOW models using the PyTorch framework.

\subsection{Package Overview}

This section outlines the primary RiverText packages and classes. Further information can be found in the accompanying API documentation, which provides a more comprehensive explanation.

\subsubsection{\texttt{utils}} The \texttt{utils} package implements utility classes and functions for code execution. The main classes are: 

\begin{itemize}
    \item \texttt{Vocab}: Container class that stores the words used as vocabulary from a text stream. This vocabulary implementation is used for all incremental word embedding algorithms.
    \item \texttt{TweetStream}: It is an iterable class that extends the \texttt{IterableDataset} class of the Pytorch API. This class allows for the efficient reading and loading of larger text files that may not be stored in memory.
\end{itemize}

\subsubsection{{\texttt{models}}} The package \texttt{models} implement the algorithms and the class for the incremental word embedding methods, considering the two approaches mentioned in subsection \ref{ssec:inc}.

\begin{itemize}
    \item \texttt{IWVBase}:  The base class represents a general interface for implementing incremental word embedding methods. This class extends the \texttt{Transformer} and \texttt{VectorizeMixin} classes of the \texttt{river} library. The \texttt{Transformer} class contains the \texttt{learn\_one} and \texttt{learn\_many} methods representing the one-example-at-a-time and mini-batch procedures mentioned in the section \ref{ssec:inc}. \texttt{VectorizeMixin} contains standard functions for text preprocessing, such as tokenization. 
    \item \texttt{IWordContextMatrix}: 
The class that implements the Incremental Word Context model. This class extends the \texttt{IWVBase} class.
    \item \texttt{IWord2Vec}: The class that implements the Incremental SkipGram and Continuous Bag of Words models. This class extends the \texttt{IWVBase} class.
    \item \texttt{iword2vec\_utils}: subpackage containing classes that implement the algorithm that incrementally updates the unigram table and the preprocessing algorithms that transform the text data into positive and negative samples.
\end{itemize}

\subsubsection{\texttt{evaluator}}

The \texttt{evaluator} package contains the class and the algorithms that implement the Periodic Evaluation mentioned in the subsection \ref{ssec:pe}.

\section{Experiments and Results}\label{sec:experiments}

In this section, we present our bechmarks results divided into three subsections. In the first part,  we explain the dataset used in this work, the second part describes the experimental setup and main hyperparameters, and the last part shows our main findings.

\subsection{Data}

Our experiment uses a dataset of unlabeled tweets to simulate a text stream of tweets. Twitter provides an excellent source of text streams, given its widespread use and real-time updates from its users. We draw a set of ten million tweets in English from the Edinburgh corpus \cite{petrovic2010edinburgh}. This dataset is a collection of tweets from different languages for academic purposes and was downloaded from November 2009 to February 2010 using the Twitter API \footnote{https://developer.twitter.com/en/docs/twitter-api}. We hypothesize that using this dataset of tweets as a text stream would allow us to evaluate the performance of incremental WE methods in a realistic scenario, given the nature of social media text and its dynamic and evolving nature.

    \subsection{Experimental setup}

In our experimental investigation, we executed the Periodic Evaluation using a diverse range of datasets and hyperparameter settings. The evaluation was conducted on multiple architectural configurations (IWCM, ISG, and ICBOW) and intrinsic test datasets \cite{jastrzebski2017evaluate}. The hyperparameters under consideration were the size of the embedding, the window size, the context size, and the number of negative samples. The results of this evaluation provide valuable insights into the performance of the different architectural configurations and hyperparameter settings, offering a comprehensive understanding of the subject under examination.

For the intrinsic test datasets, we used two datasets from the similarity tasks (MEN \cite{bruni2014multimodal} and Mturk \cite{radinsky2011word}) and one from the categorization task (AP \cite{almuhareb2005concept}). 

\subsubsection{Hyperparameter settings}

The main hyperparameter configurations that we studied were:

\begin{enumerate}
    \item We evaluated the impact of three hyperparameters on neural network embedding:
    \begin{itemize}
        \item Embedding size: refers to the dimensionality of the vector representation associated with each vocabulary word. Our configurations considered three different embedding sizes, including 100, 200, and 300.
        \item Window size: This refers to the number of neighboring tokens used as the context for a target token. Our configurations utilized three different window sizes, including 1, 2, and 3.
        \item The number of Negative samples: This refers to the number of negative instances that maximize the probability of a word being in the context of a target word. Our configurations considered three different numbers of negative samples, including 6, 8, and 10.
    \end{itemize}

Therefore, our experimental investigation considered a total of 27 configurations, comprising all combinations of the hyperparameters ($emb\_size \in {100, 200, 300}$, $window\_size \in {1, 2, 3}$, and $num\_ns \in {6, 8, 10}$) and for each of the architectural configurations and intrinsic test datasets.

    \item For the word context matrix embedding:
        \begin{itemize}
            \item We leveraged the same configurations of the embedding size and window size as we did for the neural network embedding
            \item Context size: represents the number of words associated with a vocabulary word based on the distributional hypothesis. The study involved three different context sizes, including 500, 750, and 1000. 
        \end{itemize}
        Therefore, 27 configurations were executed, incorporating all the possible combinations of ($emb\_size \in {100, 200, 300}$, $window\_size \in {1, 2, 3}$, and $context\_size \in {500, 750, 1000}$) for each intrinsic test dataset. 

It is important to mention that the vocabulary size in all configurations was set to capture 1,000,000 words. Additionally, the period value, p, utilized in our experiments was set to 320,000 instances, with a batch size of 32. This period value was selected as it represents the point at which the evaluator was called after processing 320,000 tweets. These parameters were carefully selected in order to effectively analyze the performance of the different incremental word embedding models.
\end{enumerate}

\subsection{Results}

Our analysis thoroughly evaluated various configurations (243 in total), considering the combination of three architectures, multiple hyperparameter values, and intrinsic evaluation tasks. As an illustration, we present two examples of the executed configurations:

\begin{itemize}
\item A configuration comprised of the ICBOW model, with hyperparameters set to $emb\_size = 300$, $window\_size = 3$, and $num\_ns = 10$, was employed for the similarity evaluation task using the MEN dataset.
\item Another configuration involved the ISG model, with hyperparameters defined as $emb\_size = 100$, $window\_size = 1$, and $num\_ns = 6$, was utilized for the categorization evaluation task using the AP dataset.
\end{itemize}

To determine the optimal hyperparameter configuration for each architecture and across all tasks, we employed a ranking system based on the democratic voting procedure, Borda Count \cite{emerson2013original}. The steps involved in this ranking system are as follows:

\begin{itemize} 
\item First, the mean value of each hyperparameter configuration and test dataset is computed based on the results obtained from the time series analysis. 
\item Secondly, for each test dataset that was evaluated, the average mean value is calculated across all intrinsic tasks.
\item Finally, we ordered the obtained average, with the lower position indicating the optimal configuration. 
\end{itemize}

By employing this ranking system, we aim to analyze the best hyperparameter configurations for each model and test dataset, considering that the intrinsic tasks' results are unrelated.

In Table \ref{table:rank}, we illustrate each model's top three ranked hyperparameter configurations. The complete ranking, including all configurations, can be found on the documentation page\footnote{https://dccuchile.github.io/rivertext/benchmark/}. It is crucial to mention that while the example Table showcases the best three configurations for each model, the full ranking encompasses a broader range of results.

\begin{table*}[h]
  \centering
  \caption{
The Overall Ranking of the benchmark results is based on the average of the Periodic Evaluation applied across the text stream. The result tasks are calculated by finding the mean of the evaluation, and the overall mean determined by taking the average of these result tasks. This overall mean then determines the position in the ranking.}
  \begin{tabular}{lc|cccc|ccc|c}
    \toprule
    \multicolumn{2}{c}{} &
 \multicolumn{4}{|c|}{Hyperparameters} & \multicolumn{3}{|c|}{Result tasks} \\ \hline
Position & Model & Emb. size & Win. size & Num. N.S & Context size & Mean MEN & Mean Mturk & Mean AP & Overall mean \\ \hline
    \midrule
1        & ICBOW & 100       & 3         & 6        & -            & 0.488    & 0.439      & 0.294   & 0.407        \\ \hline
2        & ICBOW & 300       & 3         & 8        & -            & 0.507    & 0.428      & 0.284   & 0.406        \\ \hline
3        & ICBOW & 300       & 3         & 6        & -            & 0.508    & 0.416      & 0.289   & 0.404        \\ \hline
4       & ISG   & 100       & 1         & 8        & -            & 0.44     & 0.4        & 0.321   & 0.387        \\ \hline
5       & ISG   & 100       & 1         & 6        & -            & 0.443    & 0.393      & 0.312   & 0.383        \\ \hline
6       & ISG   & 100       & 2         & 10       & -            & 0.421    & 0.399      & 0.309   & 0.376        \\ \hline
7       & IWCM  & 100       & 3         & -        & 1000         & 0.44     & 0.343      & 0.319   & 0.367        \\ \hline
8       & IWCM  & 200       & 3         & -        & 1000         & 0.438    & 0.351      & 0.307   & 0.366        \\ \hline
9       & IWCM  & 300       & 3         & -        & 1000         & 0.439    & 0.35       & 0.307   & 0.365        \\ \hline
    
    \bottomrule
  \end{tabular}
  
  \label{table:rank}

\end{table*}

\begin{figure*}[h]
  \centering
  \includegraphics[width=\linewidth]{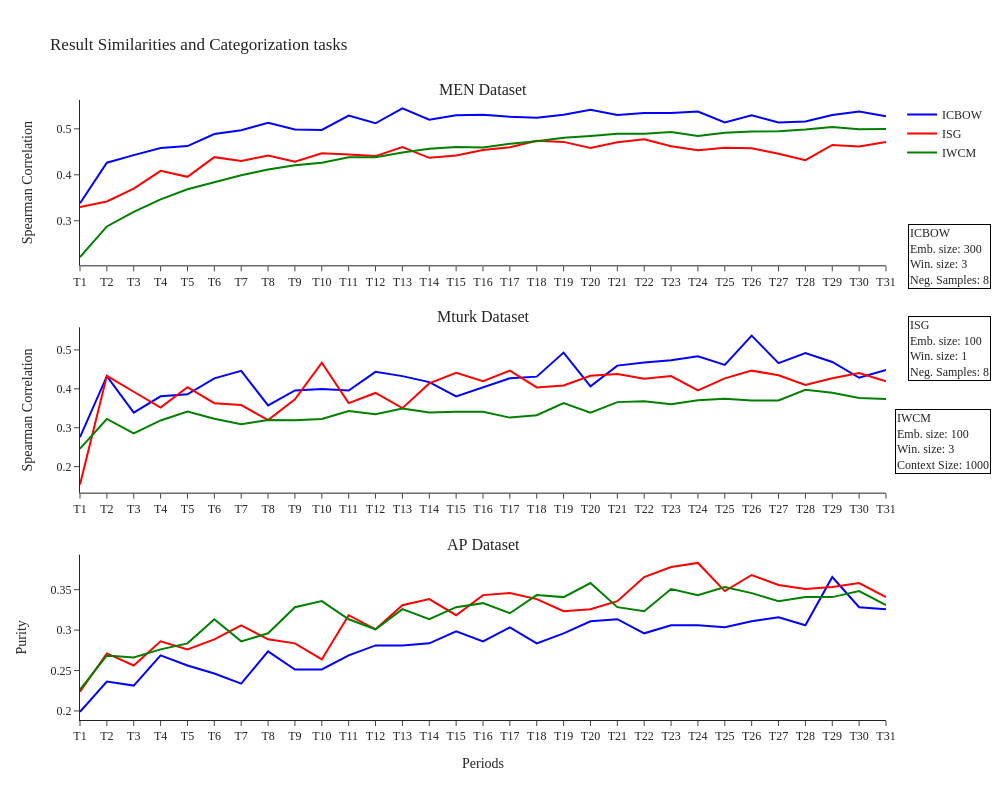}
  \caption{Best setting models for MEN, Mturk, and AP datasets. The period $p$ was set as 3,200,000 instances, which means the evaluator of the period evaluation was applied every 3,200,000 training instances.}
  \Description{}
  \label{fig:best}
\end{figure*}

As can be seen from the results, the neural network models ICBOW and ISG demonstrate superior performance, on average, compared to the non-neural network IWCM model. It is noteworthy that the ICBOW models attain better results with larger embedding and window sizes. In contrast, the ISG models perform optimally with smaller embedding and window sizes. In the case of the IWCM model, the effect of embedding and window sizes on performance is unclear. However, a trend towards improved performance with larger context sizes is observable.

Considering these findings in the context of the chosen evaluation metrics and the intrinsic tasks involved is important. The results suggest that the neural network architecture of the ICBOW and ISG models may significantly impact the performance, particularly concerning capturing semantic relationships between words. Additionally, the varying optimal configurations for the ICBOW and ISG models highlight the need for thorough experimentation and analysis when selecting hyperparameters in these models. Further research may also consider exploring the underlying mechanisms and reasons for the observed performance differences between the models.

According to Table \ref{table:rank}, we can state that the best hyperparameter setting for each model are:  

\begin{itemize}
    \item Best setting configuration for ICBOW model is $emb\_size = 100$, $window\_size = 3$, and $num\_ns = 6$.
    \item Best setting configuration for ISG model is $emb\_size = 100$, $window\_size = 1$, and $num\_ns = 8$.
    \item Best setting configuration for IWCM model is $emb\_size = 100$, $window\_size = 3$, and $context\_size = 1000$.
\end{itemize}

The results of the optimal hyperparameter configurations for each model are displayed in Figure \ref{fig:best}. This figure showcases the performance dynamics across different periods for the MEN, Mturk, and AP datasets. The figure highlights the crucial role of hyperparameter tuning in optimizing the performance of each model. The ICBOW model appears to exhibit a higher performance in the similarity task than the categorization task. Conversely, the ISG and IWCM models demonstrate better performance in the categorization task and outperform the ICBOW model. However, it is noteworthy that the results of a model for a specific intrinsic evaluation task and dataset can vary significantly and are not always related. Thus, a model may perform well in one task but poorly in another.

\section{Conclusions and Future Work}\label{sec:conc}



How humans communicate is constantly changing, amplified with the emergence of social media \cite{cunha2011analyzing}.  Existing WE methods cannot capture this dynamism in a language unless they are constantly retrained, which is computationally expensive.  Research in incremental WE has been conducted to address these limitations, yet these techniques remain disconnected from each other and lack a unified structure that would allow their comparison. Furthermore, until now, no software tool existed for easy training, evaluating, and comparing them.

In this work, we have taken a significant step by standardizing and adapting various incremental word embedding techniques \cite{bravo2022incremental, kaji2017incremental}, into a unified software framework named RiverText.  All source code is openly released and integrated into River \cite{montiel2021river}, a Python machine-learning library for data streams. 


We have conducted extensive experimentation to evaluate three incremental words embedding WE architectures, namely IWCM, ISG, and ICBOW, by varying the configuration of their hyperparameters.  Our evaluation adapts intrinsic approaches to WE evaluation, such as word similarity and categorization \cite{zhai2016intrinsic} to a streaming setting by performing periodic evaluations. Overall, our results indicate that all three architectures are competitive after appropriately tuning their hyperparameters.

For future work, we will add more incremental text representation methods, such as incremental Glove \cite{peng2018incremental}. We will also propose an evaluation methodology that incorporates concept drift, which in our case would correspond to semantic changes of words \cite{almuhareb2005concept}.  Our current periodic evaluation approach assumes that 
golden relations (e.g., word pair similarities, categories) remain static during the stream, which is inadequate for determining the abilities of our word vectors to adapt to change \cite{almuhareb2005concept}. We plan to generalize the idea developed in \cite{bravo2022incremental} to simulate synthetic tweets with semantic change for sentiment analysis to all intrinsic tasks of WE evaluation.

Additionally, we intend to enhance the functionality of our software by incorporating other sketching techniques, as outlined in previous studies such as in \cite{goyal2010sketching}, to efficiently update the vocabulary with minimal memory usage. Furthermore, we aim to integrate incremental detection of collocations or phrases, in line with recent research such as in \cite{henry2018vector}, for the representation of multi-word expressions, such as "``New Zealand'' or ``New York'', within our vocabulary.

We also plan to extend our system's capabilities by implementing a data loader to connect to the Twitter API and stream topic-specific tweets for training. This enhancement will allow users to monitor social media, providing a vast source of text streams and efficiently extracting relevant information. This functionality will expand the system's potential, enabling it to analyze the most recent and relevant data.

We hope this resource motivates further investigation in the realm of information retrieval and incremental learning, particularly concerning the representation of queries and documents in the context of time-evolving text, as seen on social media and the web.

\begin{acks}
This work was supported by ANID FONDECYT grant 11200290, the National Center for Artificial Intelligence CENIA FB210017, and ANID-Millennium Science Initiative Program - Code ICN17\_002.
\end{acks}

\bibliographystyle{ACM-Reference-Format}
\balance
\bibliography{rivertext}
\vfill\eject


\end{document}